%% file: paper.tex
\newcommand{\CLASSINPUTbaselinestretch}{1}
\documentclass[journal]{IEEEtran}
\usepackage{etoolbox}
\makeatletter
\patchcmd{\maketitle}{\@copyrightspace}{}{}{}

\makeatother
\usepackage{xspace}
\usepackage{siunitx}
\usepackage{graphicx}
\usepackage{stfloats}  
\usepackage{cite}  
\usepackage{psfrag}  
\usepackage{subfigure}
\usepackage{wrapfig}
\usepackage{epsfig}
\usepackage{url}
\usepackage{amsmath, amssymb}
\usepackage{array}
\usepackage{multicol}
\usepackage{algorithm}
\usepackage{algorithmic}
\usepackage{mathrsfs}
\usepackage{multirow}
\usepackage[normalem]{ulem}
\usepackage{verbatim}
\usepackage{fixltx2e}
\usepackage[utf8x]{inputenc} 
\usepackage{ucs} 
\usepackage{amsfonts} 
\usepackage{makeidx} 

\usepackage[justification=centering]{caption}
\usepackage{pifont}
\newcommand{\name}{X-Former}%
\usepackage{tabu}
\usepackage{floatflt}
\usepackage{amsmath, amssymb}
\usepackage{tikz}
\usepackage{multicol}
\usepackage{amsmath,amssymb}
\usepackage{lipsum,booktabs}
\usepackage{hyperref}
\usepackage{soul}

\begin{document}

\title{
  \name: In-Memory Acceleration of Transformers}
\author{\IEEEauthorblockN{Shrihari Sridharan, Jacob R. Stevens\textsuperscript{*}, Kaushik Roy and Anand Raghunathan}
\IEEEauthorblockA{\\School of Electrical and Computer Engineering, Purdue University, West Lafayette, IN, USA\\ }}


\sloppy

\maketitle
\begingroup\renewcommand\thefootnote{\textsection}
\footnotetext{\textsuperscript{*} Jacob R. Stevens is currently a Research Scientist at Meta Platforms, Inc., Cambridge, MA } 
\input{sections/abstract}


\section{Introduction}
\label{sec:introduction}
\input{sections/Introduction}
\vspace*{-0pt}

\vspace*{4pt}
\section{Preliminaries}
\label{sec:preliminaries}
\input{sections/Preliminaries.tex}

\vspace*{-0pt}

\vspace*{4pt}
\section{Related Work }
\label{sec:relatedWork}
\input{sections/relatedWork}

\vspace*{4pt}
\section{Overall Architecture}
\label{sec:tool}
\input{sections/architecture}

\vspace*{4pt}
\section{Dataflow}
\label{sec:dataflow}
\input{sections/dataflow}

\vspace*{4pt}
\section{Experimental Methodology}

\label{sec:exptsetup}
\input{sections/exptsetup}

\vspace*{0pt}
\section{Results}
\label{sec:results}
\input{sections/results.tex}


\vspace*{4pt}
\section{Conclusion}
\label{sec:conclusion}
\input{sections/conclusion}

\section{Acknowledgment}
This work was supported by C-BRIC, one of six centers in JUMP, a Semiconductor Research Corporation (SRC) program, sponsored by DARPA.
\vspace*{-0pt}
\scriptsize
\bibliographystyle{unsrt}
\bibliography{paper}

\end{document}

%% file: sections/abstract.tex
\begin{abstract}
Transformers have achieved great success in a wide variety of natural language processing (NLP) tasks due to the attention mechanism, which assigns an importance score for every word relative to other words in a sequence. However, these models are very large, often reaching hundreds of billions of parameters, and therefore require a large number of DRAM accesses. Hence, traditional deep neural network (DNN) accelerators such as GPUs and TPUs face limitations in processing Transformers efficiently. 
In-memory accelerators based on non-volatile memory promise to be an effective solution to this challenge, since they provide high storage density while performing massively parallel matrix vector multiplications within memory arrays. However, attention score computations, which are frequently used in Transformers (unlike CNNs and RNNs), require matrix vector multiplications (MVM)  where both operands change dynamically for each input. As a result, conventional NVM-based accelerators incur high write latency and write energy when used 
for Transformers, and further suffer from the low endurance of most NVM technologies.

To address these challenges, we present \name, a hybrid in-memory hardware accelerator that consists of both NVM and CMOS processing elements to execute transformer workloads efficiently. To improve the hardware utilization of \name, we also propose a sequence blocking dataflow, which overlaps the computations of the two processing elements and reduces execution time. Across several benchmarks, we show that \name{} achieves upto 85x and 7.5x improvements in latency and energy over a NVIDIA GeForce GTX 1060 GPU and upto 10.7x and 4.6x improvements in latency and energy over a state-of-the-art in-memory NVM accelerator.  
\end{abstract}

%% file: sections/Introduction.tex
{\noindent} Deep learning models that process language such as Transformers (BERT, GPT-3) have attained state of the art accuracies in a wide variety of natural language processing (NLP) tasks~\cite{bert,gpt}. These models are able to achieve superior performance due to the attention mechanism~\cite{attention}, which provides an estimate of the significance of each word with respect to every other word in a particular sequence. However, as language tasks become more and more complex, the model sizes continue to increase rapidly. For example, GPT-3 contains 175 billion parameters and acquires state of the art results for text generation with input sequence length of 2048~\cite{gpt}. Further, the attention mechanism also comprises of unique operations such as matrix vector multiplications between dynamically changing operands for every new batch of input sequences, absent in Convolutional neural networks (CNNs) and Recurrent neural networks (RNNs). Typically, on-chip scratchpads in commodity hardware are very small to store the model weights and other intermediate activations~\cite{tpu}. This leads to significant overheads due to data movement costs from off-chip accesses. For example, the energy per access of a 64-byte word from DRAM is approximately three orders of magnitude higher than on-chip access~\cite{naveen}. As a result, there are several challenges associated with implementing Transformers on existing deep learning CMOS accelerators such as GPUs and TPUs~\cite{tpu}. Consequently, in-memory computing has emerged as a viable candidate to reduce this performance bottleneck. In this paradigm, the computations are performed within the memory array itself which minimizes the number of off-chip accesses considerably. Non-volatile memory (NVM)-based in-memory compute primitives (resistive crossbars)~\cite{pcm,rram} are well suited for this application because they provide high storage density and facilitate massively parallel matrix vector multiplication (MVM) operations. However, the issues with NVM-based in-memory computing are twofold; First, Transformers require significant number of writes into the NVM device. This is a challenge due to NVMs requiring atleast one to two orders of magnitude higher latency and energy/bit depending on the NVM device compared to SRAM for the same technology node~\cite{panther,nano,over_cross}. NVM devices also have very limited endurance, only about $10^{6}-10^{9}$ conservative writes~\cite{panther,nano,over_cross} which degrades the lifetime of these devices swiftly and limits their applicability to Transformers. To alleviate the aforementioned challenges, we propose a hybrid in-memory computing architecture that combines NVM and CMOS-based processing elements to cater to the different computational and memory demands of Transformers. Our overall architecture consists of a Projection engine with NVM tiles and an Attention engine comprising of CMOS tiles, capable of executing all the operations in a transformer network efficiently. The advantage of this approach is that we can store these large models and perform the matrix multiplication operations that require no reprogramming of the NVM device in the Projection Engine. We make use of the Attention Engine to execute all the other operations since they provide high endurance, low write latency and low write energy. We also evaluate a traditional weight stationary dataflow for our architecture. However, we observe that the hardware utilization is low because the operations between different layers of a transformer encoder are dependent on one another and the size of intermediate activations scale quadratically with sequence length. Therefore, we introduce a sequence blocking dataflow which divides the input sequence into smaller blocks and processes them in a pipelined fashion. As shown in~\cite{noloss}, this technique causes no loss in application-level accuracy. Moreover, the proposed dataflow also prevents the intermediate activation sizes to scale with sequence length while keeping both the engines fully utilized. We finally compare the performance of our hardware design with a NVIDIA GeForce GTX 1060 GPU and a previously proposed crossbar-based machine learning accelerator on several benchmarks and report the improvements in energy and latency. In summary, our key contributions are:

\begin{itemize}
\item We design~\name, an in-memory hardware architecture for Transformers comprising of Projection engine and Attention engine, to efficiently accelerate attention layers in Transformers.
\item We propose an intra-layer sequence blocking dataflow for the proposed architecture, which increases the hardware utilization and reduces intermediate on-chip memory requirements.
\item We also develop a simulation framework to evaluate the performance of \name{}, and observe 85x and 7.5x improvements in latency and energy on average compared to a NVIDIA GeForce GTX 1060 GPU and upto 10.7x and 4.6x improvements in latency and energy over a state-of-the-art NVM accelerator.
\end{itemize}

The paper is organized as follows. Section II provides an overview of transformer models, its associated computational challenges and a brief background on resistive crossbars. Section III discusses the existing efforts related to this work. Section IV describes our proposed hardware architecture in detail. Section V highlights different dataflows for \name{} and lists their advantages. Section VI explains the experimental methodology used for the work. We present the performance results of \name{} in Section VII and conclude the paper in Section VIII.

%% file: sections/Preliminaries.tex
 
In this section, we describe the operations in a transformer network and analyze the various computational challenges. We also present a background on resistive crossbars.

\subsection{Transformers}
\label{subsec:DNNtrain}
\begin{figure*}[htb]
  \vspace*{-0pt}
  \centering
  \includegraphics[width=1\textwidth]{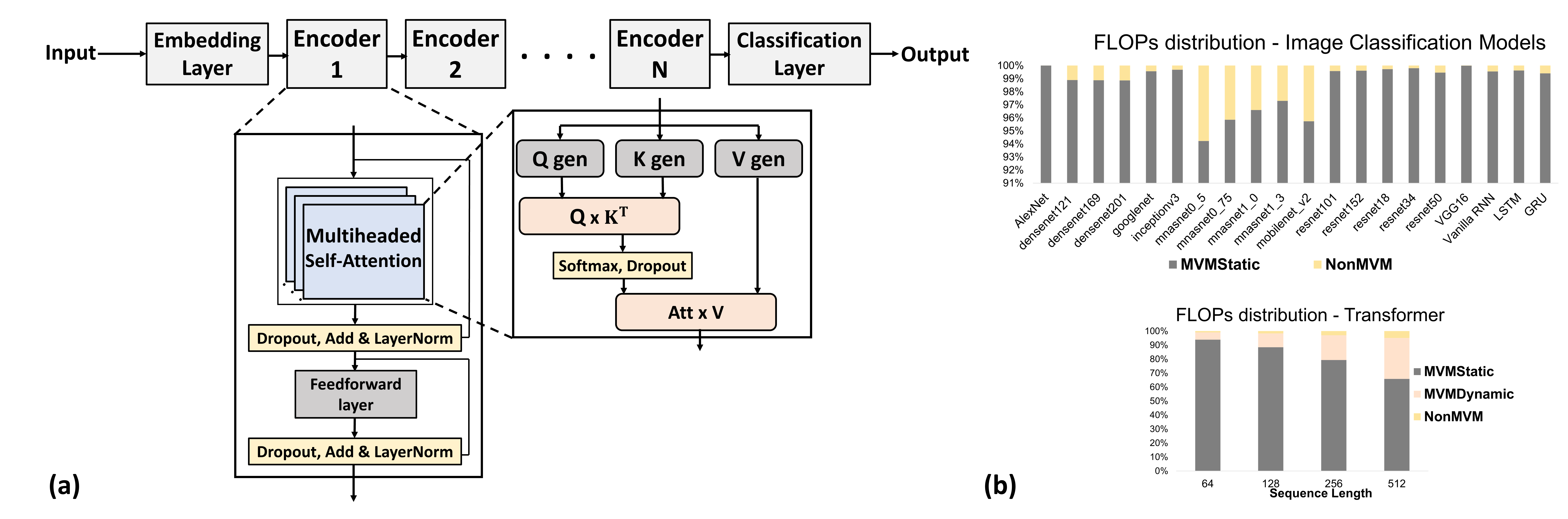}
  \vspace*{-2pt}
  \caption{(a) A transformer network with multiple encoder layers and a classification layer for a specific downstream task, (b) Distribution of Floating Point Operations (FLOPs) for image classification models and Transformers with varying sequence lengths (The distribution for image classification models is shown from 90\%-100\%)}
  \label{fig:basicarch}
  \vspace*{-8pt}
\end{figure*}

\noindent\textbf{Background.} Figure~\ref{fig:basicarch}(a) represents an example Transformer network architecture. It consists of an embedding layer followed by multiple Transformer encoders and a final classification layer fine-tuned for a specific downstream task. Every Transformer encoder is further divided into an attention layer that contains multiple self-attention heads followed by feed forward layers. Multi-headed attention layers help capture fine-grained relationships in a specific context for every word in a sequence~\cite{attention}. A batch of input sequences are first processed by an embedding layer to obtain a learned vector representation of each word in the sequence. Within each self-attention head, these inputs are multiplied by distinct weight matrices to obtain the Query, Key and Value matrices. The outputs, Query and Key matrices are now multiplied and normalized by a softmax layer to generate the intermediate attention probabilities. The intermediate attention probabilities are then multiplied with the Value matrix to obtain the final attention output. Multiple attention heads are processed in parallel and the resultant matrices are concatenated before advancing the output to the feed forward layers. This procedure is repeated for all the encoder layers.

\begin{figure}[htb]
    \centering
    \vspace*{-6pt} 
    \includegraphics[width=0.7\columnwidth]{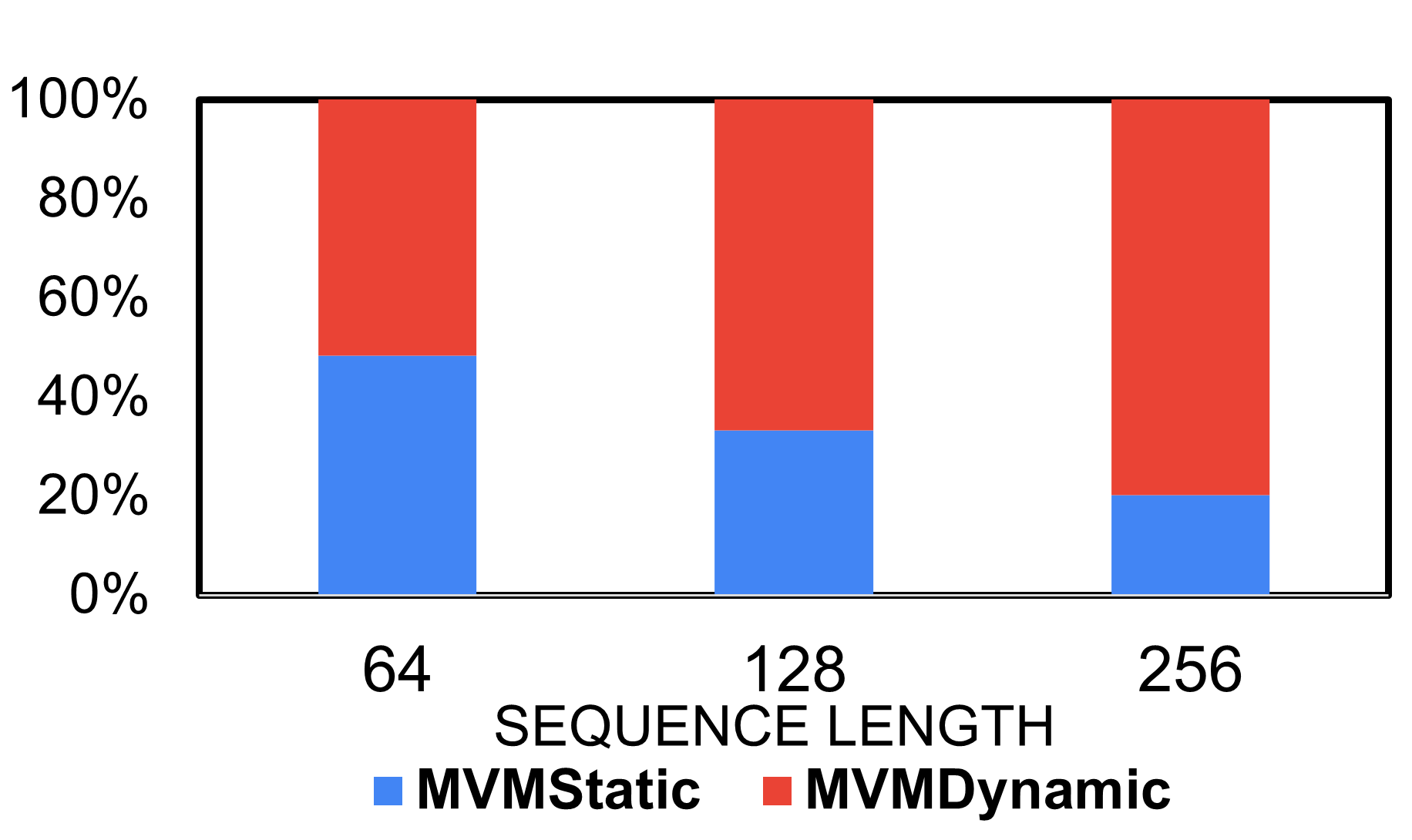}
    \caption{Runtime split of transformer operations on a traditional crossbar architecture ($MVMDynamic$ implemented in 1-D SIMD lanes)}
	\vspace*{-6pt}
    \label{fig:vfu}
\end{figure}

\noindent\textbf{Computational challenges.} Compute kernels in a Transformer network differ from traditional image classification and previously proposed language models. Matrix vector multiplication operations are typically performed between inputs/activations and weights in traditional deep learning networks. Since the weights do not require reprogramming, these operations are matrix vector multiplications with static matrices or $MVMStatic$. Transformers, in addition to $MVMStatic$ operations, also encompass matrix vector multiplications between Query, Key and Value. These data structures change for every input, hence they are dynamic and we refer to them as $MVMDynamic$ operations. We categorize other compute kernels such as softmax, layerNorm, vector operations, etc. as $NonMVM$ operations. Figure~\ref{fig:basicarch}(b) shows the floating point operations (FLOPs) distribution of various traditional deep learning networks compared to a Transformer self-attention block of varying sequence lengths. For traditional deep learning models, we can observe that $MVMStatic$ operations contributes to 95\% of the total FLOPs. However, in the case of Transformers, $MVMStatic$ operations contributes 65\% while $MVMDynamic$ operations contributes 35\% for a sequence length of 512, accounting for a substantial proportion of the overall distribution. Figure ~\ref{fig:basicarch}(b) illustrates that the FLOPs count for $MVMDynamic$ operations becomes increasingly dominant as sequence length increases. Besides, modern Transformer networks demonstrate state of the art accuracy on a variety of NLP tasks for extremely long sequences~\cite{gpt}. In traditional crossbar based architectures, this is a bottleneck because, for every self-attention layer within a transformer encoder, all the crossbars need to be reprogrammed before the matrix vector multiplication operation is performed since the inputs change dynamically. This not only causes a huge degradation in overall latency and energy but also affects the lifetime of the NVM devices, since they have very limited endurance~\cite{panther,nano,over_cross}. It is also expensive to implement these operations using the temporal 1-D single-instruction, multiple data (SIMD) lanes available in traditional in-memory accelerators. As shown in Figure ~\ref{fig:vfu}, for a sequence length of 256, 80\% of the total runtime is spent in processing $MVMDynamic$ operations. Therefore, it is very important to design a hardware architecture that can store the model parameters effectively while also improve the runtime and efficiency of $MVMDynamic$ operations for better overall performance.

%
\subsection{Resistive Crossbars}
\label{subsec:RCA}
\begin{figure}[htb]
    \centering
    \vspace*{-2pt} 
    \includegraphics[width=1\columnwidth]{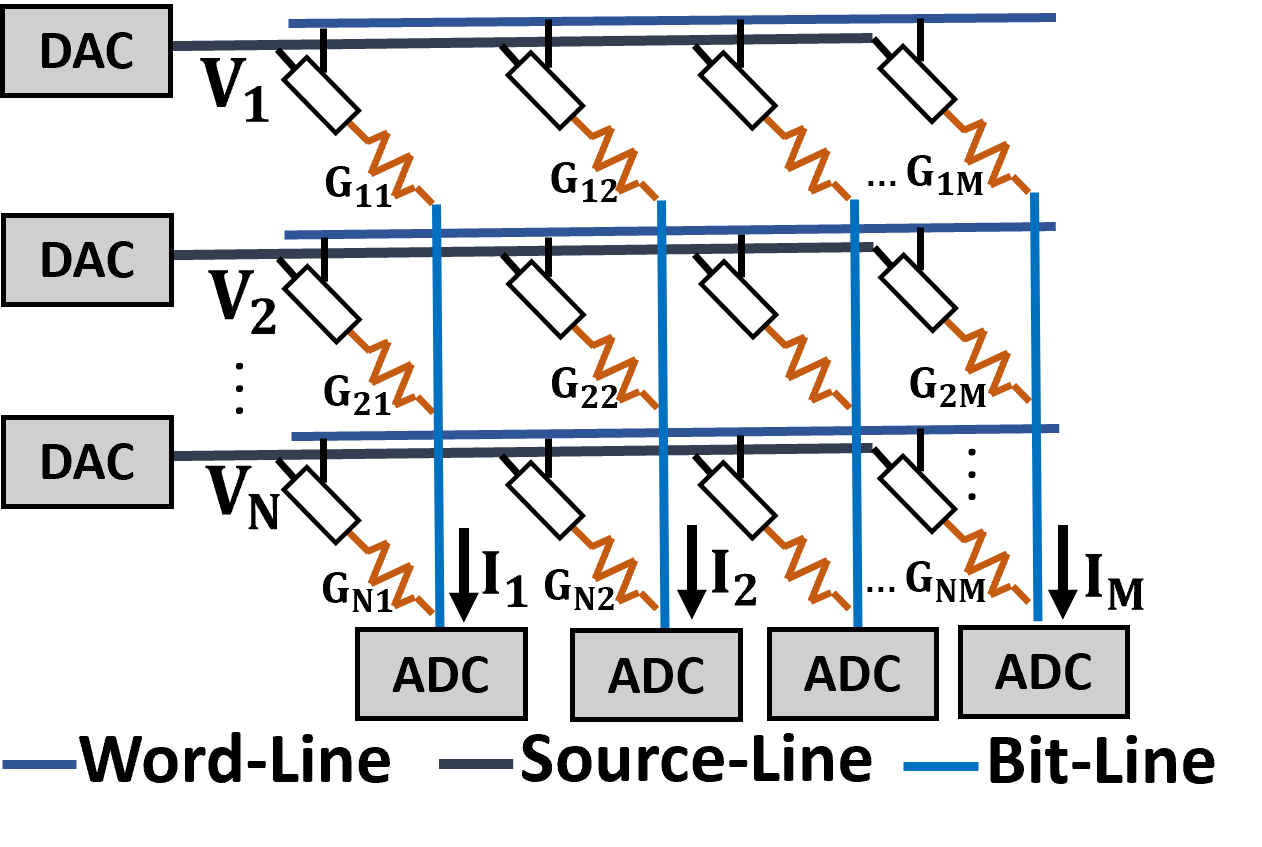}
    \caption{Structure of a resistive crossbar array with Digital to Analog converters (DACs) and Analog to Digital converters (ADCs)}
	\vspace*{-2pt}
    \label{fig:crossbar}
\end{figure}

Figure~\ref{fig:crossbar} illustrates the fundamental organization of a resistive crossbar with N rows and M columns where the NVM devices are arranged like a two dimensional array and programmed to discrete conductance states. The two terminals of the NVM device are connected via an access device to a word-line and a bit-line at each crosspoint. The digital inputs are converted to voltages using digital to analog converters (DACs). The voltages simultaneously activate all the wordlines and a multiplication operation is performed between the conductance (G) and voltage (V) at each NVM device by applying Ohm's law. The currents (I) from multiple NVM devices connected to a bit-line are accumulated as a consequence of Kirchhoff’s current law (KCL) and converted to digital outputs using analog to digital converters (ADCs). A matrix vector multiplication operation can be realized efficiently because the currents in multiple columns can be obtained simultaneously.

%% file: sections/relatedWork.tex
In this section, we describe the related research efforts and contrast our work with them. \\
\noindent\textbf{Deep learning accelerators.} Recently, various hardware accelerators have been proposed to improve the efficiency and performance of traditional deep learning networks~\cite{tpu,scaledeep,eyeriss,puma,isaac,pipelayer}. However, they are not optimized to process the unique computational kernels of Transformers described in section II. 

\noindent\textbf{Software techniques.} To improve the efficiency of Transformers, software methods such as quantization and pruning have helped reduce the bit-width of different data structures and compress the overall network~\cite{amrit,gobo,spatten,powerbert}. These efforts are complimentary to our work since they enable further improvements to our proposed hardware architecture. 

\noindent\textbf{Specialized hardware accelerators.} Another promising direction to improve the performance and energy efficiency of Transformers is to build specialized hardware for Transformers. To this end, research works such as~\cite{gobo,a3,optimus,spatten,softermax} accelerate key computations in the network. These hardware architectures have separate memory and computation units which result in significant communication overheads. An approach to reduce the number of off-chip accesses was proposed by EdgeBert~\cite{edgebert}, which stores a small part of the model (embedding table) in NVM devices. Our work differs from the above architectures by performing all the computations of the transformer network within the memory itself. 

\noindent\textbf{In-memory architectures for Transformers.} The existing processing in-memory architectures on accelerating Transformers are ~\cite{retrans} and ~\cite{translongseq}. Both the works use NVM crossbars to accelerate both $MVMStatic$ and $MVMDynamic$ operations. While they propose techniques to reduce latency of reprogramming the NVM crossbars, these devices offer very low endurance which can degrade their lifetime quickly. Our design avoids this issue by mapping $MVMStatic$ operations to NVM crossbars while processing the $MVMDynamic$ operations using in-memory CMOS processing elements. Nevertheless, we also compare our results against a traditional in-memory accelerator that maps $MVMDynamic$ operations to NVM crossbars and highlight the improvements.

%% file: sections/architecture.tex
\begin{figure}[htb]
    \centering
    \vspace*{-2pt} 
    \includegraphics[width=1\columnwidth]{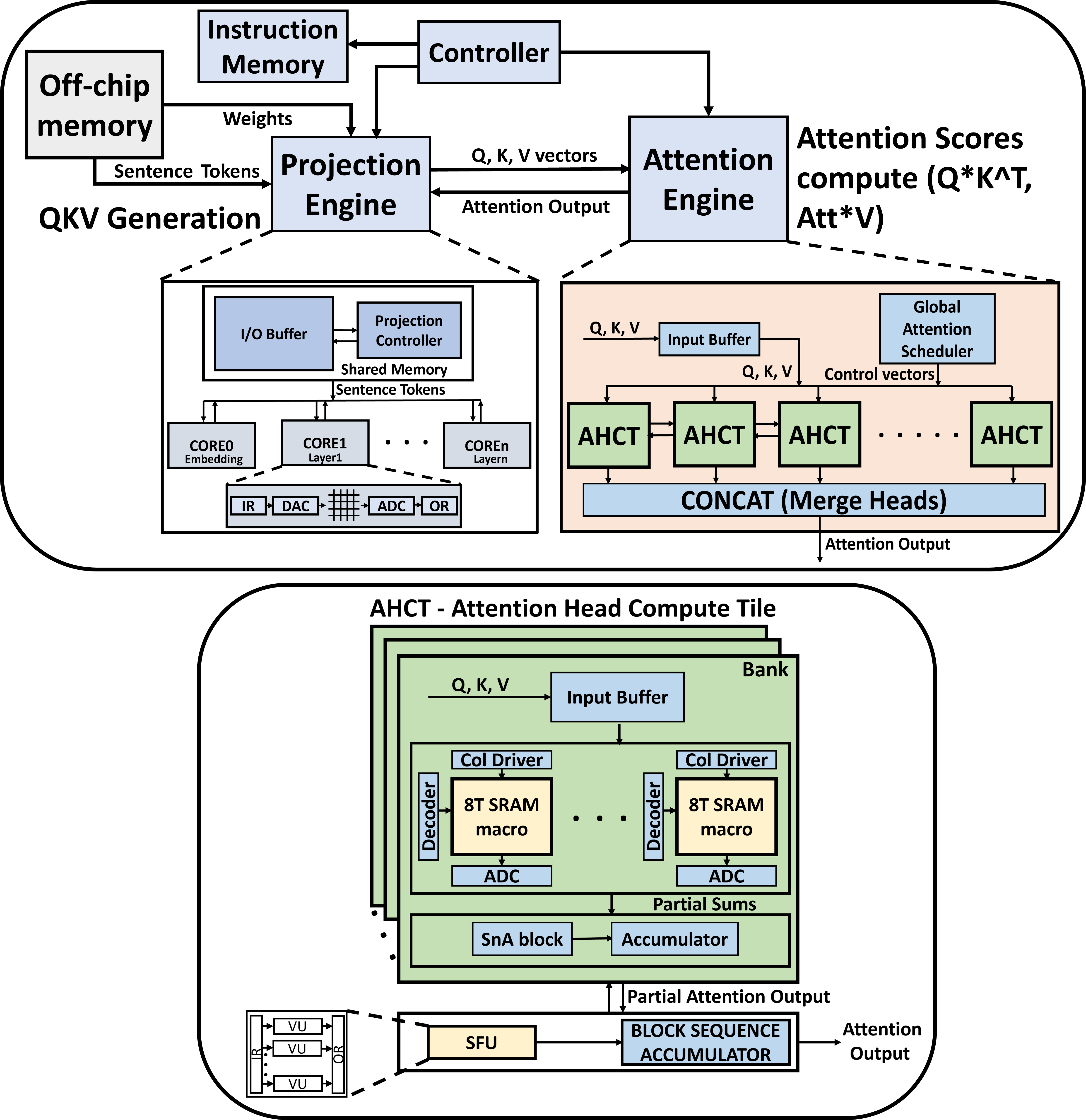}
    \caption{The \name{} hardware accelerator consisting of two different compute Engines, Projection Engine and Attention Engine and design of Attention Head Compute Tile (Bottom)}
	\vspace*{-2pt}
    \label{fig:overall_arch}
\end{figure}

\name{} is a hybrid spatial architecture that consists of two different types of processing elements to accommodate the unique needs of Transformers. We classify key workload characteristics of the transformer network into $MVMStatic$, where atleast one operand remains constant in the MVM operation for all inputs and $MVMDynamic$, where both the operands change for every batch of input sequences. \name{} is composed primarily of a Projection Engine with NVM processing tiles for executing $MVMStatic$ operations and an Attention Engine with CMOS processing tiles for executing $MVMDynamic$ operations, as shown in Figure~\ref{fig:overall_arch}. The weights of all the layers are stored in the Projection Engine to prevent reprogramming the NVM tiles, while we optimize the Attention Engine to only process the largest self-attention layer due to area constraints. We also have a dedicated bus-based interconnect network for data transfers between the Projection engine and the Attention engine and model its energy and latency. In the following subsections, each of these components will be described in more detail. 

\subsection{Projection Engine Overview}
\label{subsec:threeStage}
The Projection Engine is designed similarly to a traditional in-memory accelerator, capable of efficiently executing $MVMStatic$ operations. It consists of a hierarchical structure of tiles and cores connected via a shared memory. Each tile consists of multiple cores, and each core contains multiple resistive crossbars that can perform matrix vector multiplication operations in parallel in the analog domain. We also provision enough memory for local scratchpads to store the input and output activations within a core. The data from the scratchpads is written to the shared memory only when it is completely filled to avoid higher memory access costs. It is also important to size both the shared memory and the scratchpads appropriately to avoid diminishing the storage density advantage of resistive crossbars. The memory controller helps orchestrate and schedule the movement of data between different levels of the hierarchy. The weights are fetched from off-chip memory and programmed into the crossbars only once during compile time. Every resistive crossbar within a core is interfaced with DACs and ADCs (peripheral circuit) to convert inputs to analog values and convert the analog outputs back in the digital domain. Since the peripheral circuitry typically introduces a large overhead in area and energy, we make use of couple of techniques to reduce the cost: (i) Utilize low resolution DACs and ADCs (ii) Make multiple columns within a crossbar and multiple crossbars share both DACs and ADCs. The state of the art accuracy for Transformers inference on a variety of downstream tasks has been shown with 8-bit fixed point weights and 8-bit fixed point activations~\cite{quant}. Due to limited conductance levels in the NVM device, we use bit-slicing and bit-streaming techniques to represent the higher precision weights and activations respectively. Finally, the Query, Key and Value matrices are communicated to the Attention Engine to compute the attention scores. The Projection Engine also executes other fully connected layers present in the transformer network. Note that the projection engine stores the weights across all the layers of the network spatially, thereby avoiding writing into the resistive crossbars. This is possible due to the high storage density of the NVM devices.

As indicated in Section II, in addition to static weights in the encoder layers, the transformer network also contains an embedding layer (lookup table) that converts each word in a sequence to a continuous vector representation. Therefore, we equip the Projection Engine with additional read-only tiles to store the embedding lookup table. The read-only tiles are organized like a standard memory array and consists of NVM devices for efficient storage of the embedding weights, especially useful for longer sequence lengths. They are used only once before the first encoder layer.


\subsection{Attention Engine Overview}
\label{subsec:UpdateModel}

The Attention Engine receives the Query, Key, Value matrices from the Projection Engine and executes the $MVMDynamic$ operations in the CMOS processing tiles. To implement these multiheaded attention layers, the attention engine is composed of multiple attention head compute tiles (AHCT) that are connected via a shared bus, an on-chip transposable input buffer and a global attention scheduler to distribute the Query, Key and Value inputs to the tiles. Each AHCT is further divided into multiple banks composed of multiple 8T-SRAM cells, shift and add blocks (SnA), additional scratchpads to store partial sums and intermediate outputs and a special functional unit (SFU) to implement softmax and other vector operations. We chose 8T-SRAM cells to perform the matrix vector multiplication operations due to their decoupled read and write paths, thereby ensuring that the values are stored more robustly. The global attention scheduler assigns the Query and Value matrices to their respective banks in the AHCT. The Query and Value matrices are then written by enabling the write wordlines of the 8T-SRAM cells and driving the write bitlines to high or low depending on the value to be stored. Next, the key matrix is streamed as transposed individual vectors sequentially into the Query banks, and the matrix vector multiplication operation between the Key vector and Query matrix is computed in the analog domain to obtain the intermediate attention output. We note that similar to the Projection Engine, bit-slicing and bit-streaming are used when executing multi-bit MVM operations. The partial intermediate attention outputs are shifted and added using the shift and add (SnA) block and then stored in the accumulator. The accumulated outputs across Query banks are now processed by SFU to perform the softmax operation. The SFU consists of multiple vector units (VUs) that provides logic units such as adders, multipliers, etc. to implement simple operations such as addition, division, multiplication and exponentiation. Finally, this procedure is repeated for the MVM operation between the intermediate attention output matrix and the previously written Value matrix to get the final attention output. The output matrix is communicated back to the Projection Engine to process the feed forward layers. The Block Sequence Accumulator provides hardware support for sequence blocking dataflow, explained in more detail in Section V. All the AHCTs work in parallel to produce the final attention output, and we repeat this approach for all the layers.


%% file: sections/dataflow.tex
In this section, we describe the overall dataflow of our architecture. We first model and implement Transformers using traditional weight, Query/Value stationary dataflow. Later, we also propose a new sequence blocking dataflow to overcome some of the limitations of the earlier dataflow such as low hardware utilization and high intermediate output storage requirements. We outline both the approaches in detail considering a batch size of 1. We refer to the sequence length as SL, hidden size as HS and the hidden size within a AHCT as HSS. Both HS and HSS are properties of the language model being simulated.
\begin{figure*}[htb]
  \vspace*{-0pt}
  \centering
  \includegraphics[width=1\textwidth]{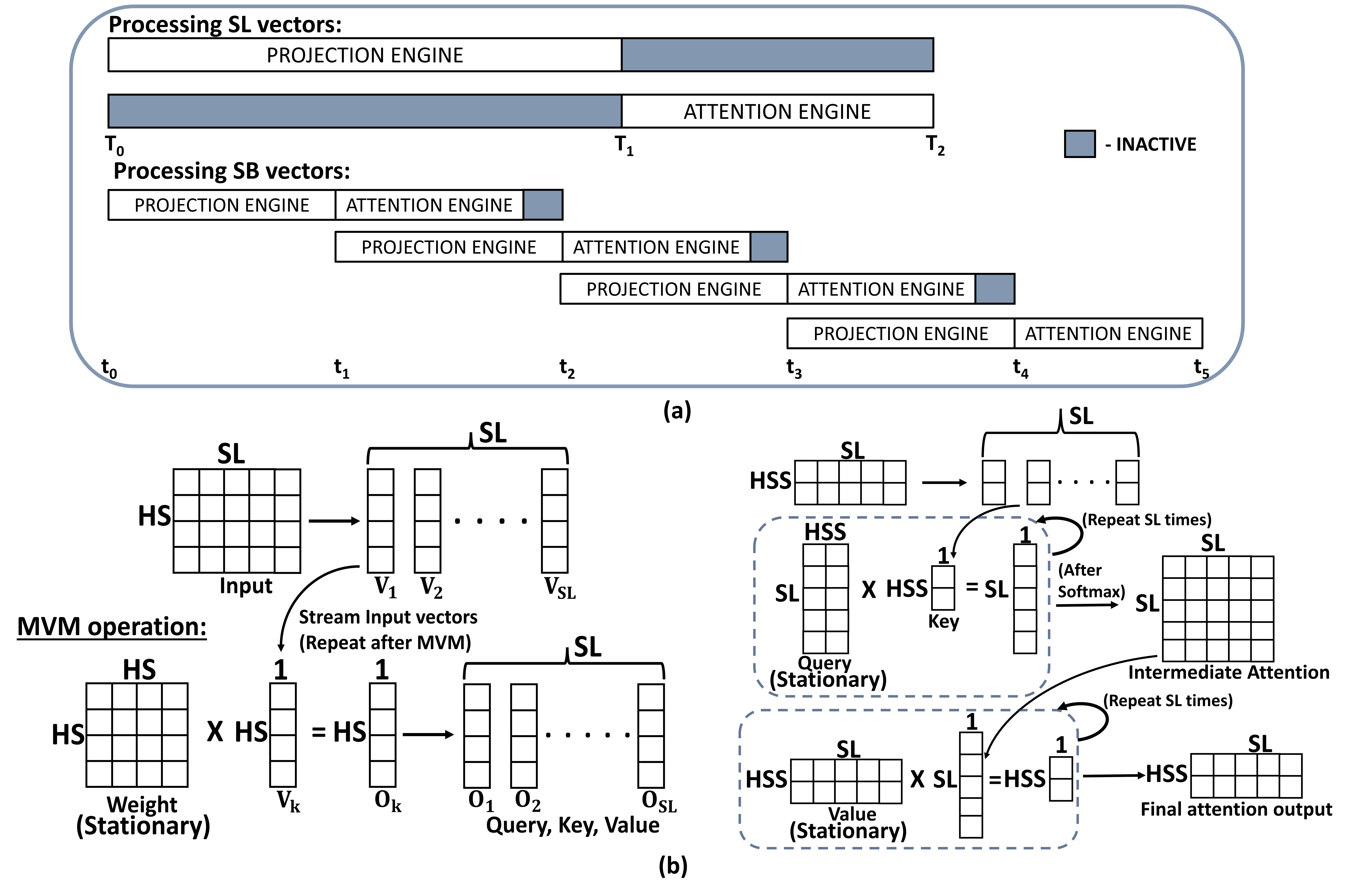}
  \vspace*{-6pt}
  \caption{(a) Active cycles of Projection Engine and Attention Engine during traditional dataflow where SL input vectors are processed and sequence blocking dataflow when SB input vectors are processed (The cycle times are not drawn to scale), (b) Weight stationary dataflow in Projection Engine (Left) and Query/Value stationary dataflow in a AHCT of the Attention Engine (Right) when SL input vectors are processed}
  \label{fig:dataflow}
  \vspace*{-6pt}
\end{figure*}

\subsection{Traditional Dataflow}
\label{subsec:DNNtrain}
\noindent\textbf{Projection Engine - Weight Stationary.} In this approach, the Projection Engine first processes an entire batch of input sequences before the Attention Engine starts computing the attention output. We adopt a weight stationary dataflow where the model weights are distributed spatially and reused across all the input sequences. Figure~\ref{fig:dataflow}(b) illustrates the process of generating the Query, Key and Value matrices for an input batch size of 1. An input sequence of size SL x HS resides in the shared memory of the Projection Engine. Once the controller delegates the cores that need to process a subset of the input, the input data is sent to the local scratchpads within a tile and in a fixed number of cycles, a vector V$_{k}$ of size SL x 1 is processed by the NVM crossbars (MVM operation) to output a vector O$_{k}$ of size HS x 1. This is repeated multiple times until the entire length of the sequence is processed to obtain Query, Key and Value matrices of size SL x HS. Owing to the storage density of NVM crossbars, the weights are distributed spatially across different tiles for these input matrices. Therefore, the matrix vector multiplication operation is parallelized for the generation of all the three matrices.


\noindent\textbf{Attention engine – Query/Value stationary.} The attention engine dataflow is described in Figure~\ref{fig:dataflow}(b). Once the Query, Key and Value matrices of size SL x HS are generated, they are split into smaller matrices of size HSS x SL and processed by each AHCT (Attention head). All the AHCTs operate in parallel and the outputs are concatenated together to obtain the final attention output. To process a HSS x SL matrix within an AHCT, we first load and write the entire Query and Value matrices into the 8T-SRAM cells of their respective banks. The write operation is performed only once for a sequence since we reuse the Query matrix for the entire sequence length. Next, we step through the set of Key vectors of size HSS x 1 SL times and aggregate the partial sums in the accumulator to obtain the intermediate attention output. Similarly, we step through the intermediate attention output vectors SL times and perform a matrix vector multiplication operation with the Value matrix to acquire the final attention output. We also observed that since the Query, Key and Value matrices are never reused outside their respective transformer encoder, the 8T-SRAM cells are overwritten for each layer. Consequently, we only provision enough SRAM tiles to process the largest layer in the network due to limited area budget. Therefore, the next encoder layer can start processing only after the current layer is completed.                                   
\subsection{Sequence blocking Dataflow}
\label{subsec:DNNtrain}
In the projection engine, we observe that every word in a sequence is handled independently to generate the Query, Key and Value matrices. We apply this fact to propose a sequence blocking dataflow, shown in Figure~\ref{fig:dataflow}(a) where a block of words in a sequence is processed together at once before the next block of words. This is not synonymous to changing the sequence length for the input because the attention score is computed for every word with respect to every other word in the overall sequence. We introduce a new parameter called sequence block (SB), which is a subset of the total sequence length. We note that multiple sequence blocks make up a sequence. 

We identified two major inefficiencies with the traditional dataflow. First, the utilization of the proposed hardware architecture is low because the Attention Engine is fully idle when the Projection Engine is computing the Query, Key, Value vectors and vice-versa. This mainly arises due to the inter-dependencies between different layers within a transformer block. With sequence blocking dataflow, once an SB length of Query, Key and Value is processed, the attention engine can start computing the attention scores. Similarly, when the attention engine is working on the previous SB, the projection engine can process the next set of inputs, keeping both the engines utilized. In addition, there is an overhead associated with storing the partial sums until the product of the length of each sequence block and the number of sequence blocks equals the length of the overall sequence. As shown in section VII, this cost is negligible in comparison to the benefits of sequence blocking.
\begin{figure}[htb]
    \centering
    \vspace*{-6pt} 
    \includegraphics[width=0.7\columnwidth]{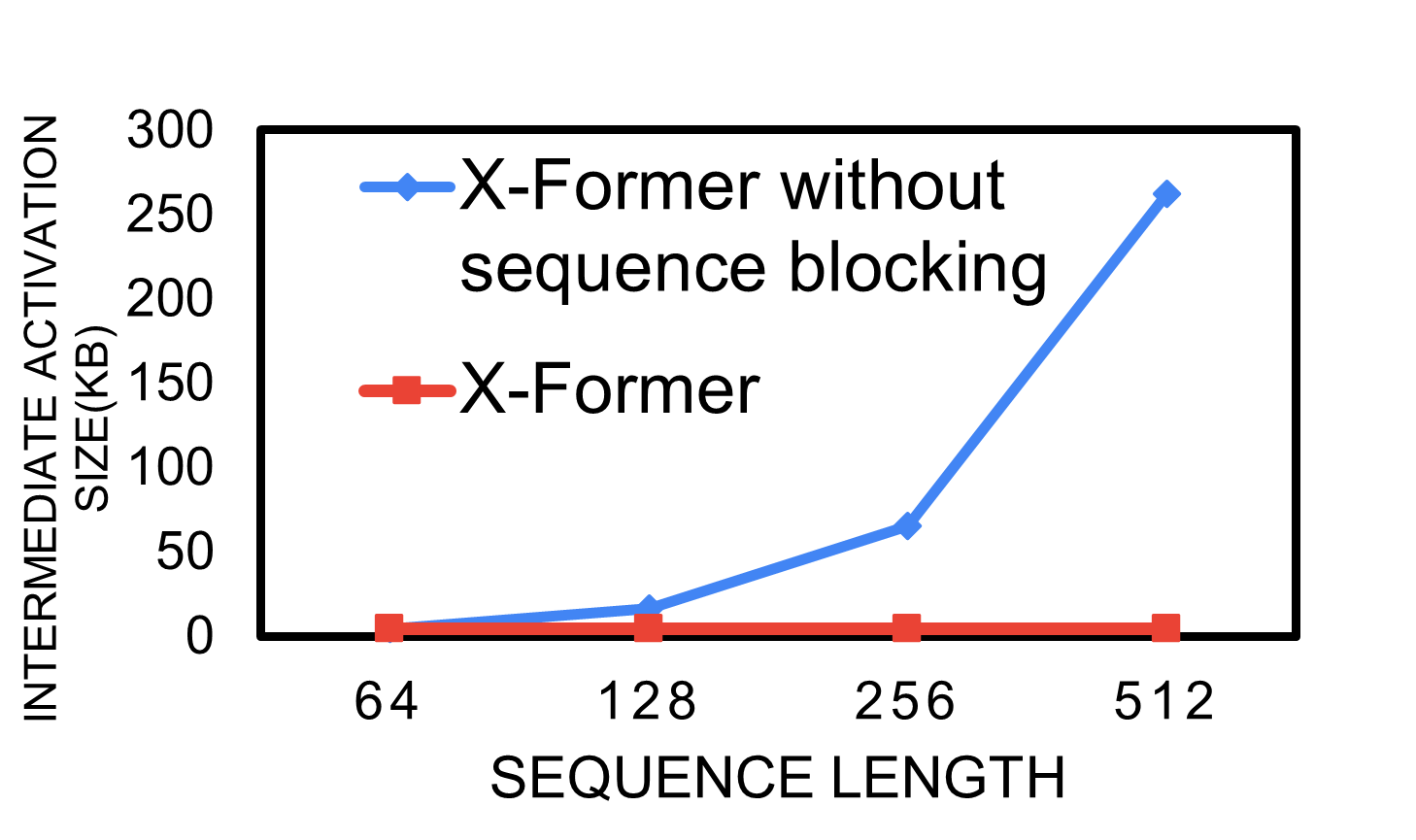}
    \caption{Intermediate activation storage requirements for increasing sequence lengths with traditional dataflow and sequence blocking dataflow}
	\vspace*{-6pt}
    \label{fig:activ}
\end{figure}

Next, since the entire sequence is processed by the Attention Engine at once, the intermediate attention output size becomes SL x SL. As the sequence length increases, this size scales quadratically, which leads to high on-chip memory requirements. In contrast, sequence blocking dataflow ensures that the intermediate attention output size (SB x SB) doesn't change, since it only depends on the sequence block length and not the total sequence length.

%% file: sections/exptsetup.tex
This section explains the experimental setup and methodology used to evaluate \name{}. \\
\noindent\textbf{System-level modeling.} We developed an architecture simulator for \name{} to model and evaluate the different operations in the transformer network. We obtained separate timing and power models for the resistive crossbar, DAC and ADC for the Projection Engine from~\cite{puma}. For the AHCT modules, we performed SPICE simulations for the MVM operations (8T-SRAM macro) to obtain power and write energy measurements. We obtained the Successive Approximation Register (SAR) ADC power measurements from \cite{adc_survey}. Other digital units were modeled as Register Transfer Level (RTL) units and synthesized using 32nm CMOS technology. We measured their power consumption using the Synopsys Design Compiler. The execution timing and power for all the on-chip memories were modeled using the CACTI simulator~\cite{cacti}. Finally, we integrated both the Projection Engine and Attention Engine into PUMAsim~\cite{puma} to obtain the execution traces and estimate the overall latency and energy. Table~\ref{tab:microarch} describes the micro-architectural configuration for \name{} that achieves the most optimal performance. We report the raw performance metrics of \name{} compared to a NVIDIA GPU in Table~\ref{tab:rawperf}.
\newcolumntype{M}[1]{>{\centering\arraybackslash}m{#1}}
\begin{table}
\caption{\name{} Micro-architectural Configuration}

\begin{center}
\begin{tabular}
{M{30mm}|M{40mm}}
\hline
\textbf{Parameter} & \textbf{Value} \\
\hline
Operating Frequency & 1 GHz \\
NVM device & ReRAM - 2bit \\
NVM Resistance &100k$\Omega$ - 1M$\Omega$ \\
Crossbar Dimension & 128x128 \\ 
DAC & 1-bit \\
ADC & 2/crossbar, 8-bit @ 1.28GS/s \\
Projection Engine Config &36 tiles; 8 cores/tile, 6 crossbars/core \\
 Shared Memory & 384KB \\ 
 Input/Output scratchpad & 4KB \\ 
 Attention Engine Config &2; 16 AHCT/PE, 8 banks/AHCT, 6 8T-SRAM cells/bank\\
 SFU & 16 VU with 4 lanes each\\
 Block Sequence Accumulator & 6KB \\
\hline

\end{tabular}
\end{center}
\label{tab:microarch}
\end{table}

\begin{table}
\caption{Summary of Network Topology}

\begin{center}
\begin{tabular}
{M{30mm}|M{20mm}|M{15mm}}
\hline
\textbf{Variable} & \textbf{BERT-base} & \textbf{BERT-large}\\
\hline
\# heads &12 &16 \\
\# layers &12 &24 \\
Hidden size (HS) &768 &1024 \\
Hidden size/head (HSS) &64 &64 \\
Bit-width (W, Q, K, V) &(8, 8, 8, 8) &(8, 8, 8, 8) \\
 
\hline

\end{tabular}
\end{center}
\label{tab:networks}
\end{table}

\begin{table}
\caption{Summary of Benchmarks}

\begin{center}
\begin{tabular}
{M{13mm}|M{35mm}|M{25mm}}
\hline
\textbf{Dataset} & \textbf{Description} &\textbf{max\_sequence\_length (SL)}\\
\hline
GLUE~\cite{glue}  & Collection of nine different language understanding tasks & 512 \\
\hline
SQuAD~\cite{squad} & Question-Answering Tasks; max\_query\_length = 64, doc\_stride = 128 & 384 \\

\hline

\end{tabular}
\end{center}
\label{tab:benchmarks}
\end{table}

\begin{table}
\caption{Performance metrics }

\begin{center}
\begin{tabular}
{M{20mm}|M{20mm}|M{15mm}|M{18mm}}
\hline
\textbf{Performance Metric} & \textbf{NVIDIA GeForce GTX 1060 GPU} &\textbf{\name{}} &\textbf{\name{} with Sequence blocking Dataflow}\\
\hline
Latency/inference  & 51.46 ms & 7.94 ms & 0.98 ms \\

TOPS/W & 0.0158 & 0.837 & 6.72 \\

\hline

\end{tabular}
\end{center}
\label{tab:rawperf}
\end{table}

\noindent\textbf{Network architectures and Benchmarks.} We evaluated \name{} on the benchmarks specified in Table~\ref{tab:benchmarks} using network configurations described in Table~\ref{tab:networks}. We note that for GLUE benchmark, the final results are an average over all the downstream tasks. In SQuAD benchmark, the doc\_stride parameter denotes the stride length when a long document is split and max\_query\_length indicates the maximum number of tokens in a question~\cite{huggingface}.


\vspace*{0pt}

%% file: sections/results.tex
\noindent In this section, we evaluate \name{} using different metrics and illustrate the benefits.

\subsection{Accuracy Evaluation}
\label{subsec:accuracy}
\begin{figure}[htb]
    \centering
    \vspace*{-6pt} 
    \includegraphics[width=\columnwidth]{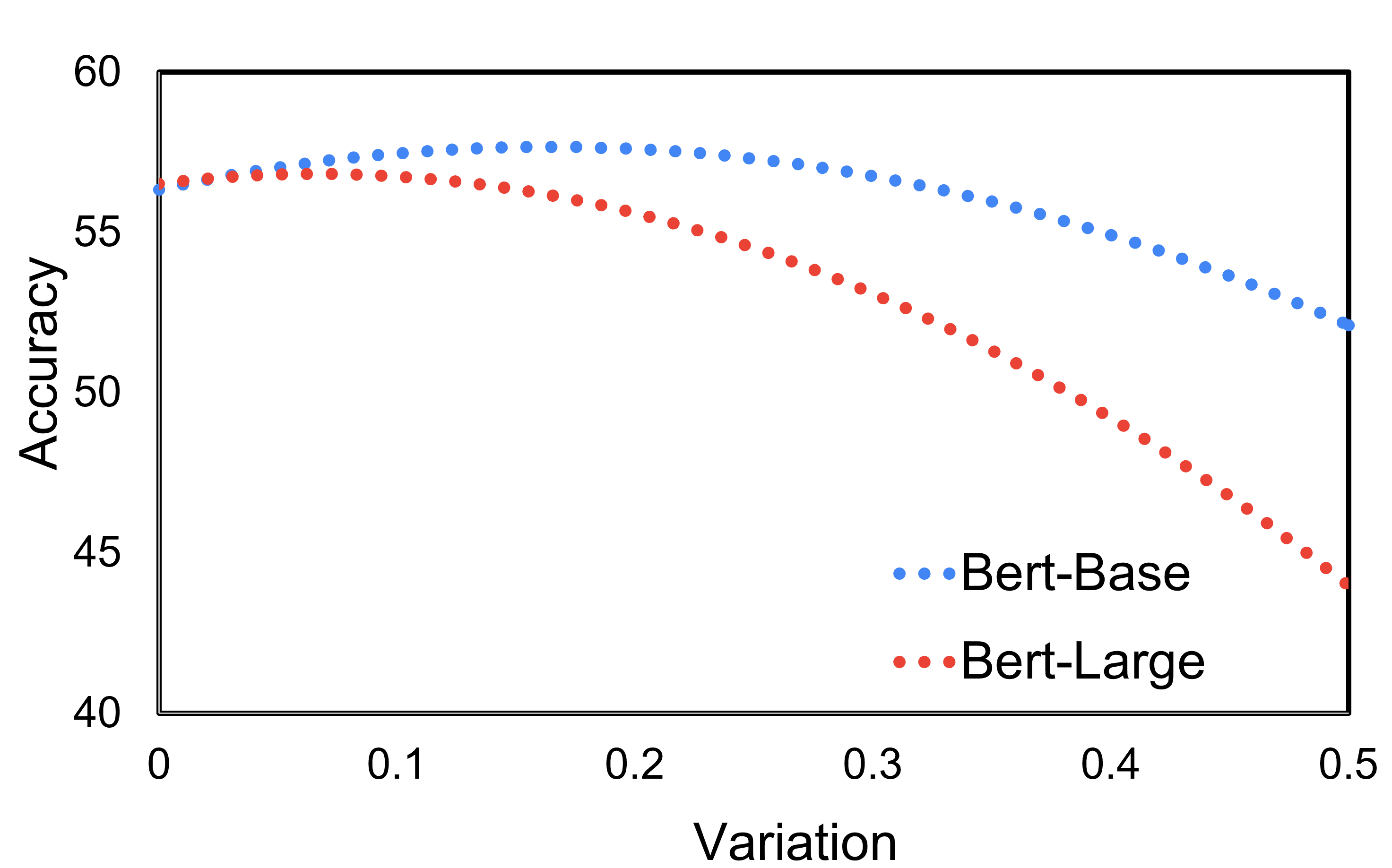}
    \caption{Application level accuracy degradation of Bert-Base and Bert-Large due to analog computations for a sequence length of 512}
	\vspace*{-6pt}
    \label{fig:acc}
\end{figure}
Due to the analog nature of computations in \name, functional errors are introduced in the inference pass that impacts the overall application level accuracy \cite{txsim, neurosim}. The major sources of errors are device and circuit non-idealities which occur during both read and write operations. These circuit non-idealities arise from source and sink resistances in the peripheral circuitry, wire parasitic resistances in the metal lines of the crossbar and sneak paths. Similarly, device non-idealities include process variations and write non-linearity. The cumulative effect of these non-idealities manifest as errors in both $MVMStatic$ and $MVMDynamic$ operation outputs. We investigated the impact of these inaccuracies for Bert-Base and Bert-Large for a sequence classification (WNLI) task under different process variations using the NeuroSim simulator ~\cite{neurosim} and summarize the results in Figure \ref{fig:acc}. The accuracy of the ideal software implementation (var=0) is 56.34 \%. At the lower device variation regime (0-0.2), we observe less than 1\% loss in accuracy. However, for larger variations, we observe approximately 4.2 \% error for Bert-Base and 9.87 \% error for Bert-Large network. The accuracy degradation of Bert-Large is higher due to errors accumulating over larger number of layers. We note that there has been active research in the devices and materials community to engineer devices with lower process variations. Moreover, this is a energy-accuracy tradeoff that has been widely studied and several mitigation strategies proposed ~\cite{cxdnn} to recover the accuracy comparable to the ideal software implementation.

\subsection{Performance Results}
\label{subsec:speed}
\begin{figure}[htb]
    \centering
    \vspace*{-6pt} 
    \includegraphics[width=\columnwidth]{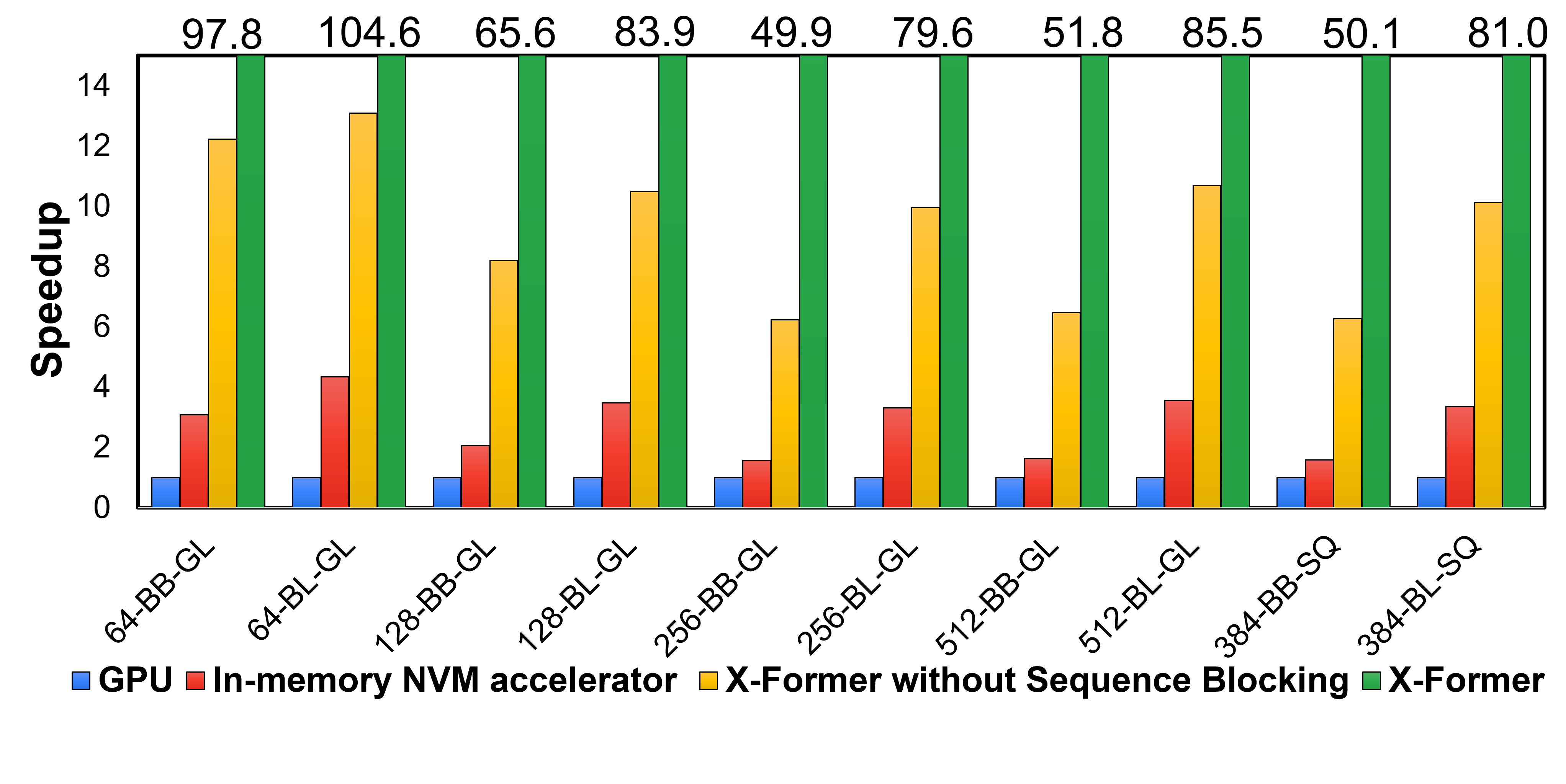}
    \caption{Normalized speedup results for varying sequence lengths for GLUE and SQUAD benchmarks.BB refers to Bert-base, BL refers to Bert-Large, GL refers to GLUE dataset and SQ refers to SQuAD dataset (Large values clipped) }
	\vspace*{-6pt}
    \label{fig:perf}
\end{figure}
Figure \ref{fig:perf} shows the normalized speedup results with reference to the execution time of a NVIDIA GeForce GTX 1060 GPU. We consider two variations of \name: (i) Without Sequence Blocking dataflow where the execution of the Projection Engine and Attention Engine do not overlap (ii) With the proposed sequence blocking (SB=64) dataflow. We also compare our results with an in-memory NVM accelerator ~\cite{puma} where the $MVMDynamic$ operations are executed in the temporal 1-D SIMD lanes instead of the NVM tiles due to limited endurance. First, without sequence blocking, we observe upto 10x benefits on average for both GLUE and SQuAD benchmarks over GPU. This is because \name{} is a highly parallel spatial in-memory architecture. Moreover, we also observe improvements compared to a traditional NVM in-memory accelerator since the Attention Engine is more efficient in processing the MVM operation compared to the 1-D SIMD module (vector-vector operations). Next, we notice additional improvements with sequence blocking, upto 85x and 81x for GLUE and SQuAD benchmarks respectively because the Attention Engine is inactive only until the Projection Engine processes a subset of the input instead of the entire sequence length. Therefore, both the engines work efficiently with higher utilization, resulting in better throughput. In addition, the size of the intermediate activations are a function of the sequence block (SB) rather than the total length of the sequence. Consequently, the size of the intermediate activations reduces significantly, eliminating the need for off-chip accesses. We note that, as shown in Figure \ref{fig:perf}, for larger sequence lengths, the speedup deteriorates because the matrix vector multiplication operation starts consuming a larger fraction of the execution time. 
\begin{figure}[htb]
    \centering
    \vspace*{-6pt} 
    \includegraphics[width=0.7\columnwidth]{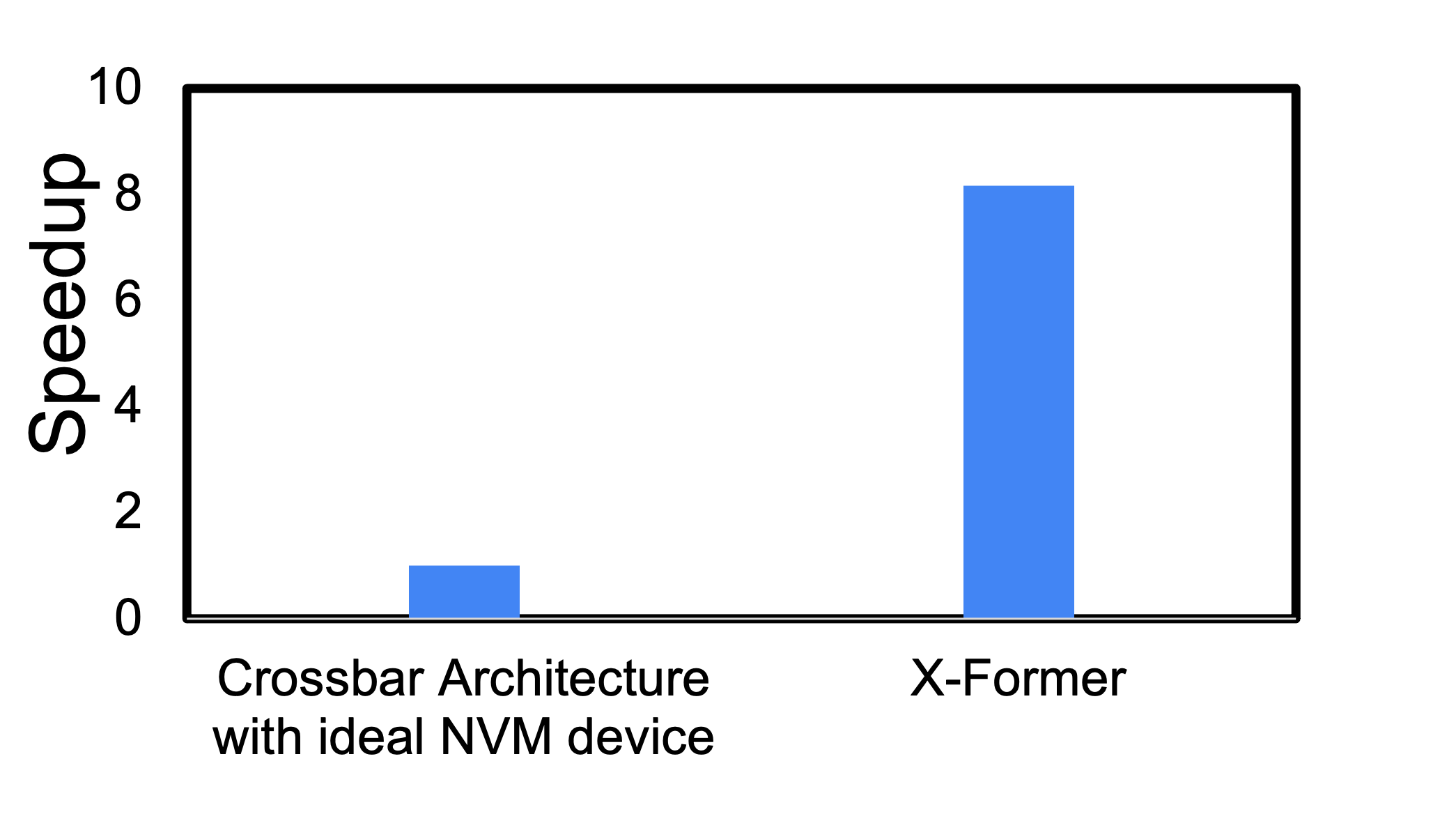}
    \caption{Performance benefits of \name{} compared to a crossbar architecture with NVM\textsubscript{Ideal} device}
	\vspace*{-6pt}
    \label{fig:hypo}
\end{figure}

We also compare \name{} with two in-memory architectures with homogeneous memory technologies and explain the advantages of our hybrid solution. First, we consider an ideal scenario where there exists an NVM device (NVM\textsubscript{Ideal}) that offers very high endurance and negligible write errors. For our evaluations, we assume an existing crossbar architecture that has the same technology node and area as \name{} and all the MVM operations mapped to the NVM tiles. As shown in Figure \ref{fig:hypo}, our architecture achieves 8.1x speedup over the NVM architecture. This is due to NVMs requiring high latency for performing the $MVMDynamic$ operations (both write and compute included). We also note that sequence blocking dataflow cannot be implemented in the crossbar architecture with NVM\textsubscript{Ideal} device since the Attention Engine requires additional execution time due to the write latency, which increases the inactivity of the Projection Engine and results in under-utilization of the hardware. 
\begin{figure}[htb]
    \centering
    \vspace*{-6pt} 
    \includegraphics[width=\columnwidth]{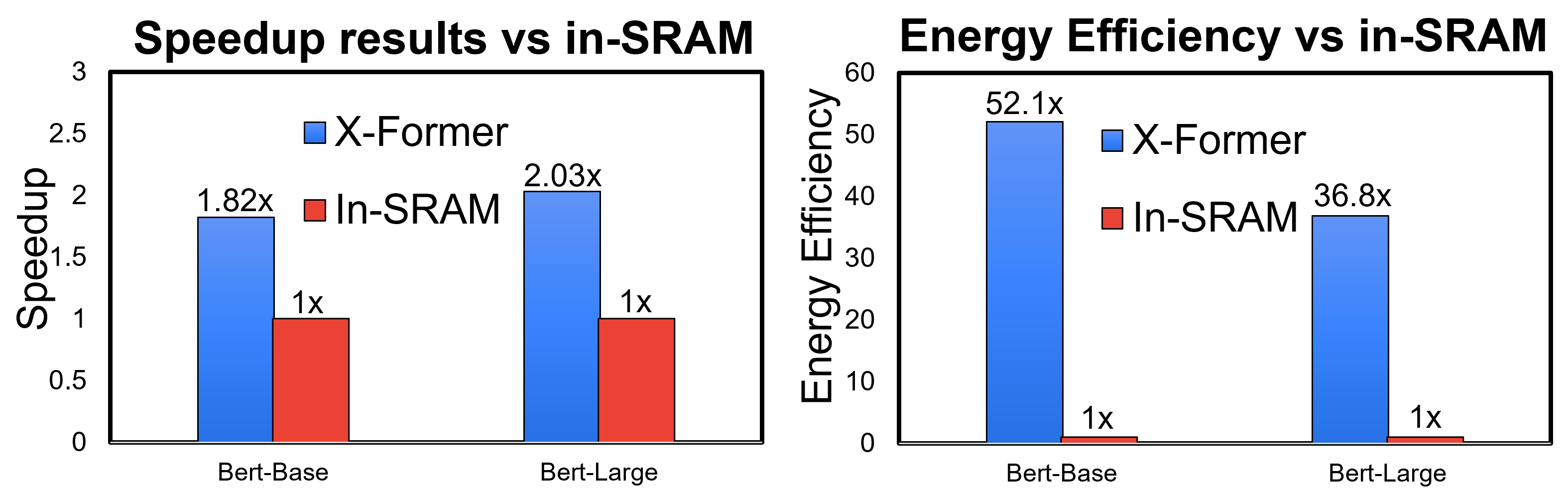}
    \caption{Normalized speedup and energy efficiency of \name{} compared to baseline in-memory SRAM architecture}
	\vspace*{-6pt}
    \label{fig:iSRAM}
\end{figure}
\par Next, we consider an in-memory SRAM setup under iso-area, where both the $MVMStatic$ and the $MVMDynamic$ operations are computed in the SRAM arrays. We note that the bit cell area of 1T-1R ReRAM is approximately 8F\textsuperscript{2} while the area of an SRAM cell is 150F\textsuperscript{2}. Even though some of this density advantage is lost due to peripherals, the effective area of ReRAM arrays is much lower than SRAM arrays. Therefore, under the iso-area baseline, an off-chip DRAM is modeled from which the weights are fetched as required. We adopt a temporal architecture design for the in-memory SRAM setup (\name{} is spatial), where weights are written into the Projection Engine and Attention Engine. Other effects such as writing to the SRAM cells are also considered. Although $MVMStatic$ and $MVMDynamic$ operations can be scheduled in both the engines due to homogeneity, most of the tiles are required to process only one type of compute. This is because the majority of the chip is required to process $MVMDynamic$ and performing the attention computation for that layer can cause overwriting of some layer weights. We also observe that only a tiny fraction of the off-chip access latency can be hidden when $MVMDynamic$ is executed since it is 4.55x slower.

In Figure ~\ref{fig:iSRAM}, we compare the speedup and energy efficiency of X-former with the in-SRAM accelerator for GLUE tasks and a sequence length of 512. We observe that X-former with the traditional dataflow is 1.82x and 2.03x for Bert-Base and Bert-Large networks respectively. This speedup is due to X-former not requiring any off-chip accesses since ReRAMs are dense and can store the weights of all the layers on chip. Similarly, the energy efficiency is 52.1x and 36.8x due to the off-chip accesses and SRAM write energy. The latency can be further optimized by adopting a sequence blocking dataflow for X-Former.

\subsection{Energy Results}
\label{subsec:speed}
\begin{figure}[htb]
    \centering
    \vspace*{-6pt} 
    \includegraphics[width=\columnwidth]{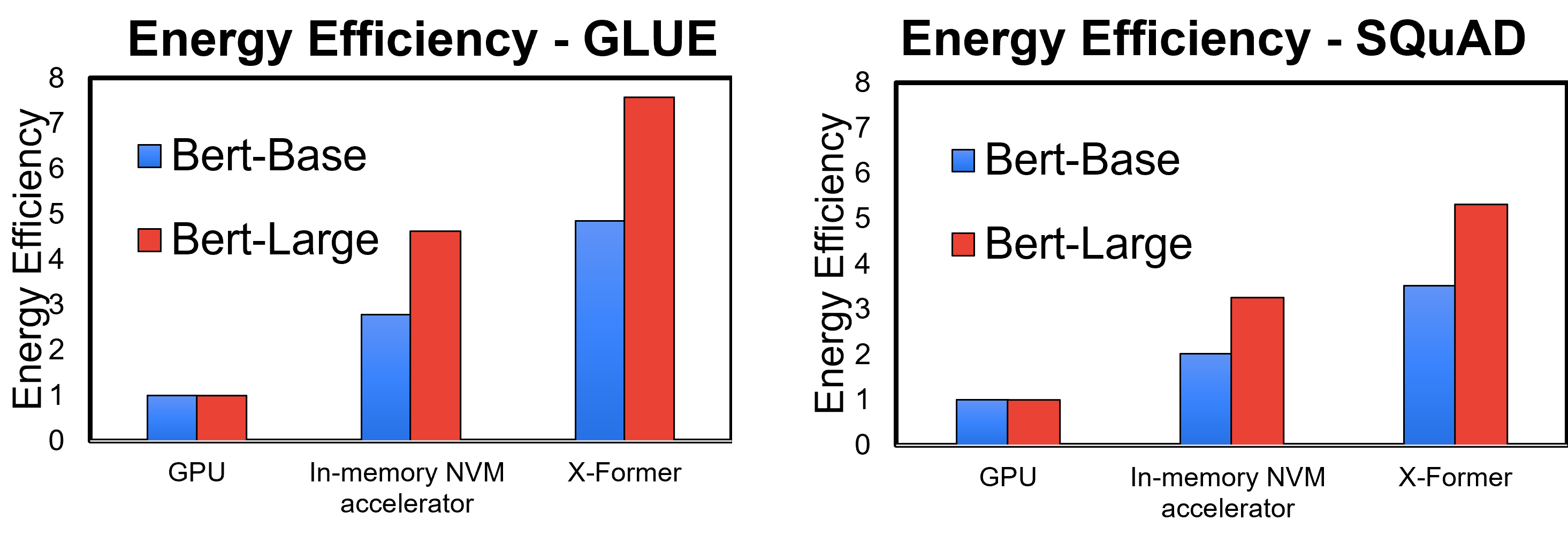}
    \caption{Normalized energy efficiency of \name{} compared to NVIDIA GPU and a state-of-the-art in-memory accelerator}
	\vspace*{-6pt}
    \label{fig:energy}
\end{figure}

Figure \ref{fig:energy} shows the normalized energy efficiency of \name{} compared to a GPU and an in-memory NVM accelerator for two different benchmarks. In order to calculate the GPU energy, we measured the average power and execution time using utility tools such as nvidia-smi. We achieve 4.85x-7.5x and 3.52x-5.31x energy benefits for GLUE and SQuAD benchmarks respectively with different configurations of the transformer network. Since NVMs provide very high storage density, the weights are already programmed in-memory leading to low data movement costs. However, GPUs require a large number of off-chip accesses, which increases its overall energy. 

\begin{figure}[htb]
    \centering
    \vspace*{-6pt} 
    \includegraphics[width=\columnwidth]{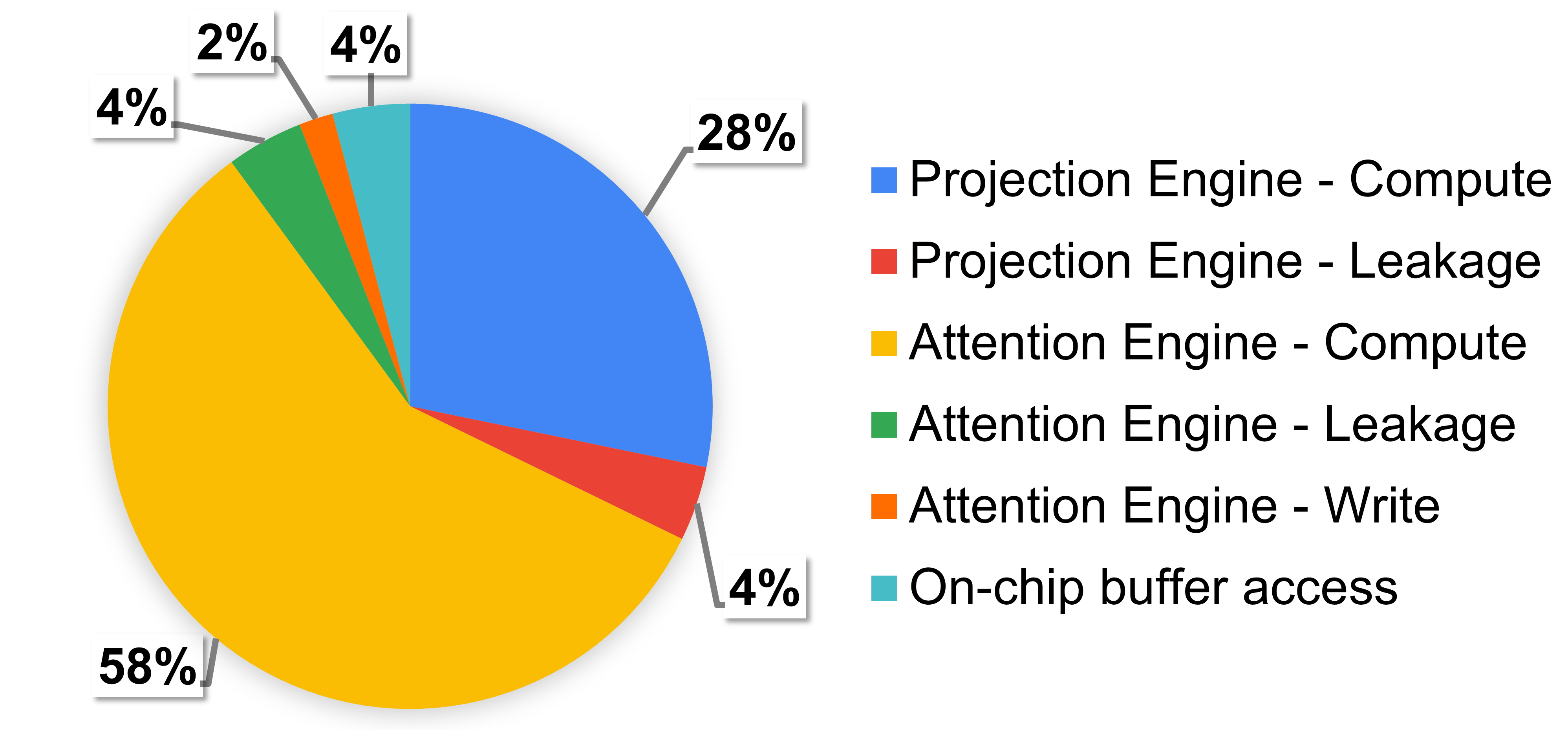}
    \caption{Energy breakdown of running a single Transformer encoder in \name{} for a sequence length of 512}
	\vspace*{-6pt}
    \label{fig:energy_break}
\end{figure}

\noindent As shown in Figure~\ref{fig:energy_break}, we report the overall energy breakdown of different components of the X-Former architecture for GLUE benchmark. We observe that, out of 15.2 $\mu$J, the majority of energy (58\%) is consumed in computing the relevance scores in the Attention Engine while 28\% is spent on generating the query, key and value outputs in the Projection Engine. Although the number of FLOPs for $MVMDynamic$ operations is lower (35\%) than $MVMStatic$ operations (65\%) for a sequence length of 512, a larger fraction of the energy is spent on computing attention scores because the dynamic power required to perform an MVM operation using a CMOS memory array is 4x more than the NVM memory array \cite{srammac}. We also observe that the attention engine write energy is only 2\% (0.27 $\mu$J) of the overall write energy since CMOS processing elements are used to process attention even though the FLOPs distribution is high. We note that the block sequence accumulator and SFU in the attention engine only contribute a very tiny fraction (0.02 \%) of the total attention engine energy.

\begin{figure}[htb]
    \centering
    \vspace*{0pt} 
    \includegraphics[width=0.8\columnwidth]{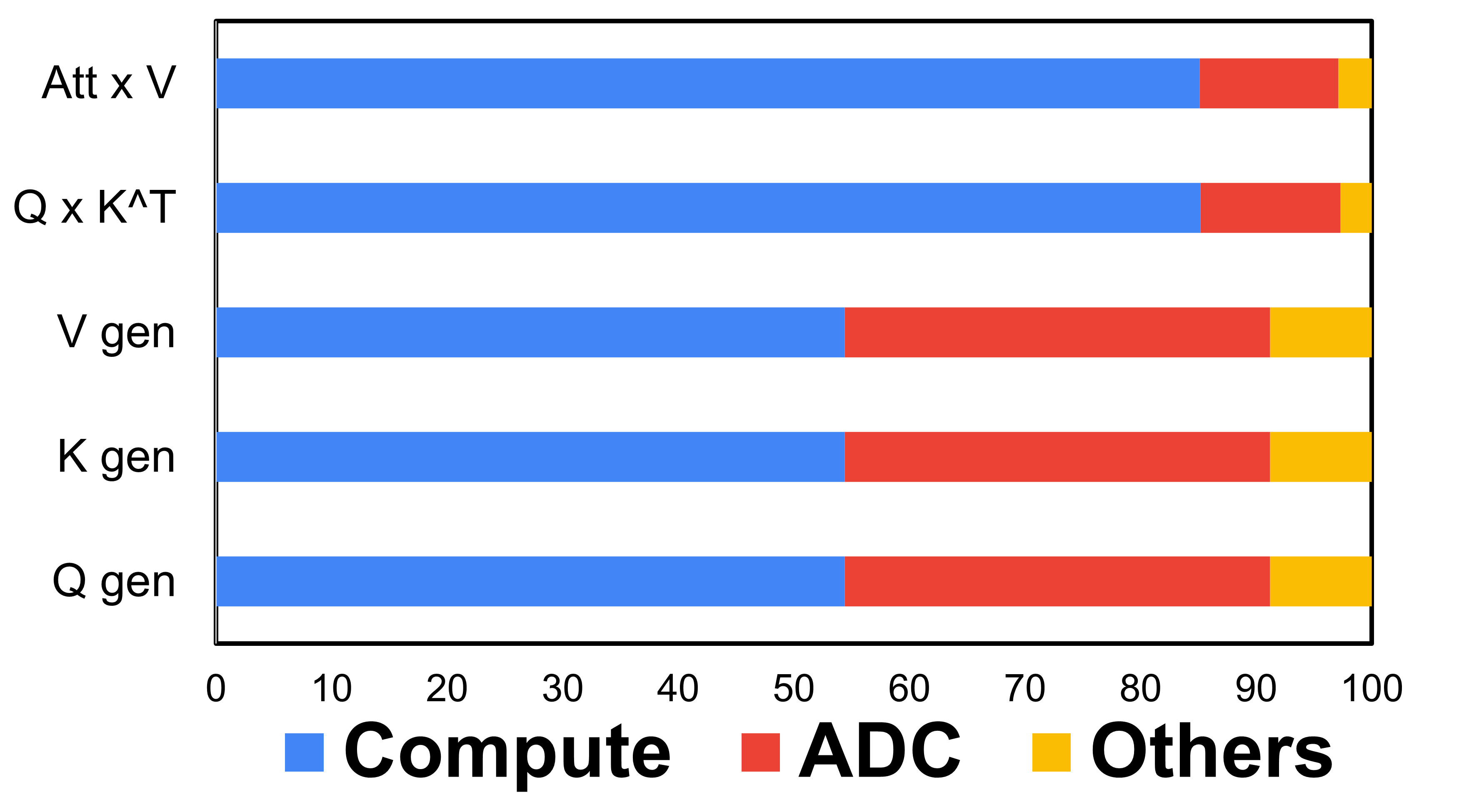}
    \caption{Energy breakdown of all MVM operations in a attention layer for a sequence length of 512}
	\vspace*{0pt}
    \label{fig:mvm_break}
\end{figure}

We also present the energy breakdown in terms of different operations within a transformer encoder as shown in Figure~\ref{fig:mvm_break}. Note that Q gen, K gen and V gen are processed by Projection Engine while Q x K$^{T}$ and Att x V layers are realized using the Attention Engine. The ADC energy is more dominant (36.8\%) in NVM processing elements while it contributes lesser to the overall MVM energy in Attention Engine (12.1\%) since the ratio of MVM compute energy to ADC energy in a CMOS memory array is higher than the NVM memory array.  

\subsection{Scaling \name{}}
\label{subsec:scale}

\name{} consists of a combination of ReRAM cells as in-memory compute (IMC) units to store model weights, ReRAM as standard memory to store embedding weights, SRAM cells as IMC units and buffers for inputs and outputs, intermediate activations and partial sums. Since we need the SRAM arrays to only store a single layer of the network, the attention engine in \name can scale to arbitrary number of layers. We would only need more ReRAM cores in the Projection Engine to store the model weights. Like most previous NVM architectures, we assume a spatial architecture, where the amount of ReRAM tiles needs to be increased to support the larger networks. The reason for this design choices the limited endurance problem of ReRAM and very high write energy costs. An alternate approach would be to scale \name{} to multiple nodes with chip-to-chip interconnects.

%% file: sections/conclusion.tex
{\noindent} Transformer networks are a promising class of machine learning workloads that have greatly improved the state of the art accuracy in a wide variety of natural language processing tasks. However, these networks possess several memory and computational challenges that prohibit traditional deep learning accelerators to realize them efficiently. To this end, we propose \name{}, a hybrid in-memory accelerator consisting of both NVM and CMOS processing tiles that can store the model weights and perform $MVMDynamic$ operations efficiently. We also propose a intra-layer sequence blocking dataflow to improve the hardware utilization and reduce the size of the intermediate activations. Our evaluations, considering different network configurations and benchmarks, reveal that \name{} can lead to significant improvements in latency and energy over a NVIDIA GeForce GTX 1060 GPU and a previously proposed state-of-the-art in-memory NVM accelerator.